\def\year{2020}\relax
\documentclass[letterpaper]{article} 
\usepackage{aaai20}  
\usepackage{times}  
\usepackage{helvet} 
\usepackage{courier}  
\usepackage[hyphens]{url}  
\usepackage{graphicx} 
\usepackage{amsmath}
\usepackage{amsfonts}
\usepackage{xcolor}
\usepackage{siunitx}
\usepackage{multirow}

\newcommand{\xinit}{\ensuremath{x_{\textit{init}}}}
\newcommand{\xfinal}{\ensuremath{x_{\textit{final}}}}
\newcommand{\xobject}{\ensuremath{x_{\textit{object}}}}
\newcommand{\xdistractor}{\ensuremath{x_{\textit{distractor}}}}

\usepackage{graphicx}  
\title{That and There: Judging the Intent of Pointing Actions with Robotic Arms}

\author{Malihe Alikhani, Baber Khalid\thanks{Equal contribution}, Rahul Shome\footnotemark[1], Chaitanya Mitash\\
\bf \Large Kostas Bekris, Matthew Stone\\
Computer Science, Rutgers University\\
110 Frelinghuysen Road\\
Piscataway, NJ 08854-8019\\
firstname.lastname@rutgers.edu
}

\begin{document}

\maketitle
\begin{abstract}
\begin{quote}

Collaborative robotics requires effective communication between a robot and a human partner. This work proposes a set of interpretive principles for how a robotic arm can use pointing actions to communicate task information to people by extending existing models from the related literature. These principles are evaluated through studies where English-speaking human subjects view animations of simulated robots instructing pick-and-place tasks. The evaluation distinguishes two classes of pointing actions that arise in pick-and-place tasks: referential pointing (identifying objects) and locating pointing (identifying locations). The study indicates that human subjects show greater flexibility in interpreting the intent of referential pointing compared to locating pointing, which needs to be more deliberate. The results also demonstrate the effects of variation in the environment and task context on the interpretation of pointing. Our corpus, experiments and design principles advance models of context, common sense reasoning and communication in embodied communication.

\end{quote}
\end{abstract}

\section{Introduction}
\label{intro}

Recent years have seen a rapid increase of robotic deployment, beyond traditional applications in cordoned-off workcells in factories, into new, more collaborative use-cases. For example, social robotics and service robotics have targeted scenarios like rehabilitation, where a robot operates in close proximity to a human. While industrial applications envision full autonomy, these collaborative scenarios involve interaction between robots and humans and require effective communication. For instance, a robot that is not able to reach an object may ask for a pick-and-place to be executed in the context of collaborative assembly. Or, in the context of a robotic assistant, a robot may ask for confirmation of a pick-and-place requested by a person.

When the robot's form permits, researchers can design such interactions using principles informed by research on embodied face-to-face human--human communication.  In particular, by realizing \emph{pointing gestures}, an articulated robotic arm with a directional end-effector can exploit a fundamental ingredient of human communication \cite{kita2003pointing}.  This has motivated roboticists to study simple pointing gestures that identify objects \cite{han2018placing,holladay2014legible,zhao2016experimental}.   This paper develops an empirically-grounded approach to robotic pointing that extends the range of physical settings, task contexts and communicative goals of robotic gestures. This is a step towards the richer and diverse interpretations that human pointing exhibits \cite{kendon:2004}.

This work has two key contributions.  First, we create a systematic dataset, involving over 7000 human judgments, where crowd workers describe their interpretation of animations of simulated robots instructing pick-and-place tasks.  Planned comparisons allow us to compare pointing actions that identify objects (referential pointing) with those that identify locations (locating pointing). They also allow us to quantify the effect of accompanying speech, task constraints and scene complexity, as well as variation in the spatial content of the scene.  This new resource documents important differences in the way pointing is interpreted in different cases.  For example, referential pointing is typically robust to the exactness of the pointing gesture, whereas locating pointing is much more sensitive and requires more deliberate pointing to ensure a correct interpretation.  The Experiment Design section explains the overall process of data collection, the power analysis for the preregistered protocol, and the content presented to subjects across conditions.

\begin{figure}[t]
    \centering
    \includegraphics[width=0.45\textwidth]{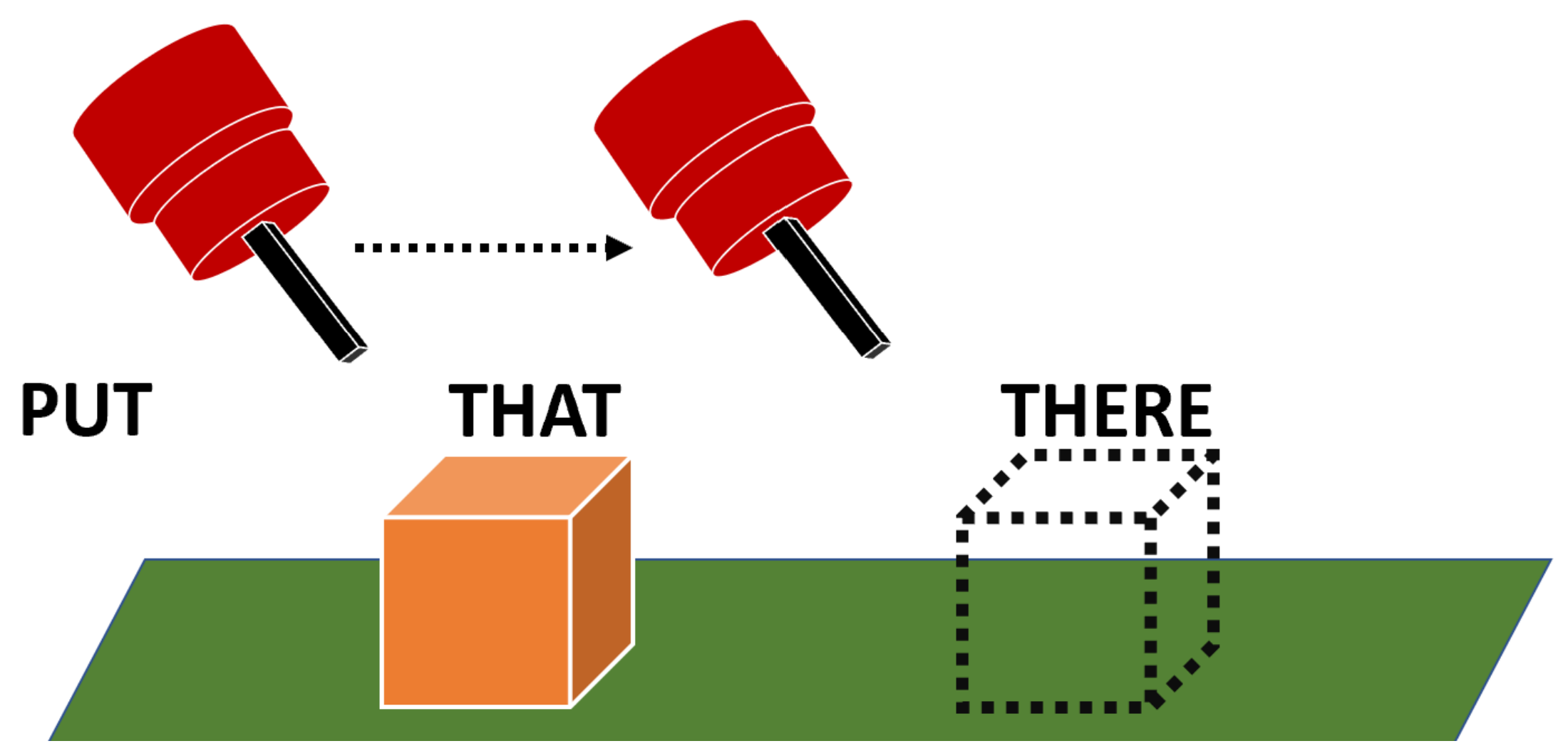}
    \caption{A pick-and-place task requires a \textit{referential} pointing action to the object (orange cube) at the initial position, and a \textit{locating} pointing action to a final placement position (dotted cube). Such an action by a robot (in red) can also be accompanied by verbal cues like \textit{``Put that there.''}}
    \label{fig:pap}
\end{figure}

The second contribution is a set of interpretive principles, inspired by the literature on vague communication, that summarize the findings about robot pointing.  They suggest that pointing selects from a set of candidate interpretations determined by the type of information specified, the possibilities presented by the scene, and the options compatible with the current task.  In particular, we propose that pointing picks out all candidates that are not significantly further from the pointing ray than the closest alternatives.  Based on our empirical results, we present design principles that formalize the relevant notions of ``available alternatives'' and ``significantly further away'', which can be used in future pointing robots.  The Analysis and Design Principles sections explain and justify this approach.

\section{Related work}
\label{related-work}

This paper focuses on the fundamental AI challenge of effective embodied communication, by proposing empirically determined generative rules for robotic pointing, including not only referential pointing but also pointing that is location-oriented in nature. Prior research has recognized the importance of effective communication by embracing the diverse modalities that AI agents can use to express information. In particular, perceiving physical actions \cite{thibadeau1986artificial} is often essential for socially-embedded behavior \cite{dautenhahn2002embodied}, as well as for understanding human demonstrations and inferring solutions that can be emulated by robots \cite{kuniyoshi1994learning}. Animated agents have long provided resources for AI researchers to experiment with models of conversational interaction including gesture \cite{cassell:siggraph1994}, while communication using hand gestures \cite{pavlovic1997visual} has played a role in supporting intelligent human-computer interaction. 

Enabling robots to understand and generate instructions to collaboratively carry out tasks with humans is an active area of research in natural language processing and human-robot interaction \cite{butepage2017human,cha2018survey}. Since robotic hardware capabilities have increased, robots are increasingly seen as a viable platform for expressing and studying behavioral models \cite{scassellati2000investigating}.  In the context of human-robot interaction, deictic or pointing gestures have been used as a form of communication \cite{pook1996deictic}.  More recent work has developed richer abilities for referring to objects by using pre-recorded, human-guided motions \cite{sauppe2014robot}, or using mixed-reality, multi-modal setups \cite{williams2019mixed}. 

Particular efforts in robotics have looked at making pointing gestures legible, adapting the process of motion planning so that robot movements are correctly understood as being directed toward the location of a particular object in space \cite{holladay2014legible,zhao2016experimental}.  The current work uses gestures, including pointing gestures and demonstrations, that are legible in this sense. It goes on to explore how precise the targeting has to be to signal an intended interpretation.

In natural language processing research, it's common to use an expanded pointing cone to describe the possible target objects for a pointing gesture, based on findings about human pointing \cite{kranstedt2003deixis,rieser2004pointing}.  Pointing cone models have also been used to model referential pointing in human--robot interaction \cite{whitney2016interpreting,whitney2017reducing}. In cluttered scenes, the pointing cone typically includes a region with many candidate referents.  Understanding and generating object references in these situations involves combining pointing with natural language descriptions \cite{han2018placing,kollar2014grounding}.  While we also find that many pointing gestures are ambiguous and can benefit from linguistic supplementation, our results challenge the assumption of a uniform pointing cone. We argue for an alternative, context-sensitive model.

In addition to gestures that identify objects, we also look at pointing gestures that identify points in space. The closest related work involves navigation tasks, where pointing can be used to discriminate direction (e.g., left vs right) \cite{mei2016listen,tellex2011understanding}. The spatial information needed for pick-and-place tasks is substantially more precise. Our findings suggest that this precision significantly impacts how pointing is interpreted and how it should be modeled.

\section{Communicating Pick-and-Place}
\label{problem}

This section provides a formalization of pick-and-place tasks and identifies information required to specify them.
 
\noindent\textbf{Manipulator}: Robots that can physically interact with their surroundings are called \textit{manipulators}, of which robotic arms are the prime example. 

\noindent\textbf{Workspace}: The manipulator operates in a 3D workspace $\mathcal{W} \subseteq \mathbb{R}^3$. The workspace also contains a stable surface of interest defined by a plane $S\subset\mathcal{W}$ along with various objects. To represent 3D coordinates of workspace positions, we use $x\in\mathcal{W}$. 

\noindent\textbf{End-effector}: The \textit{tool-tips} or \textit{end-effectors} are geometries, often attached at the end of a robotic arm, that can interact with objects in the environment. These form a manipulator's chief mode of picking and placing objects of interest and range from articulated fingers to suction cups. A subset of the workspace that the robot can \textit{reach} with its end-effector is called the reachable workspace. The end-effector in this work is used as a pointing indicator.

\noindent\textbf{Pick-and-place}: Given a target object in the workspace, a \textit{pick-and-place} task requires the object to be picked up from its initial position and orientation, and placed at a final position and orientation. When a manipulator executes this task in its reachable workspace, it uses its end-effector. 
The rest of this work ignores the effect of the object's orientation by considering objects with sufficient symmetry. Given this simplification, the pick-and-place task can be viewed as a transition from an initial position $\xinit\in\mathcal{W}$ to a final placement position $\xfinal\in\mathcal{W}$.  Thus, a pick-and-place task can be specified with a tuple
$$ \textit{PAP} = \langle o, \xinit, \xfinal \rangle. $$

\begin{figure}[t]
\centering
\includegraphics[width=0.5\textwidth]{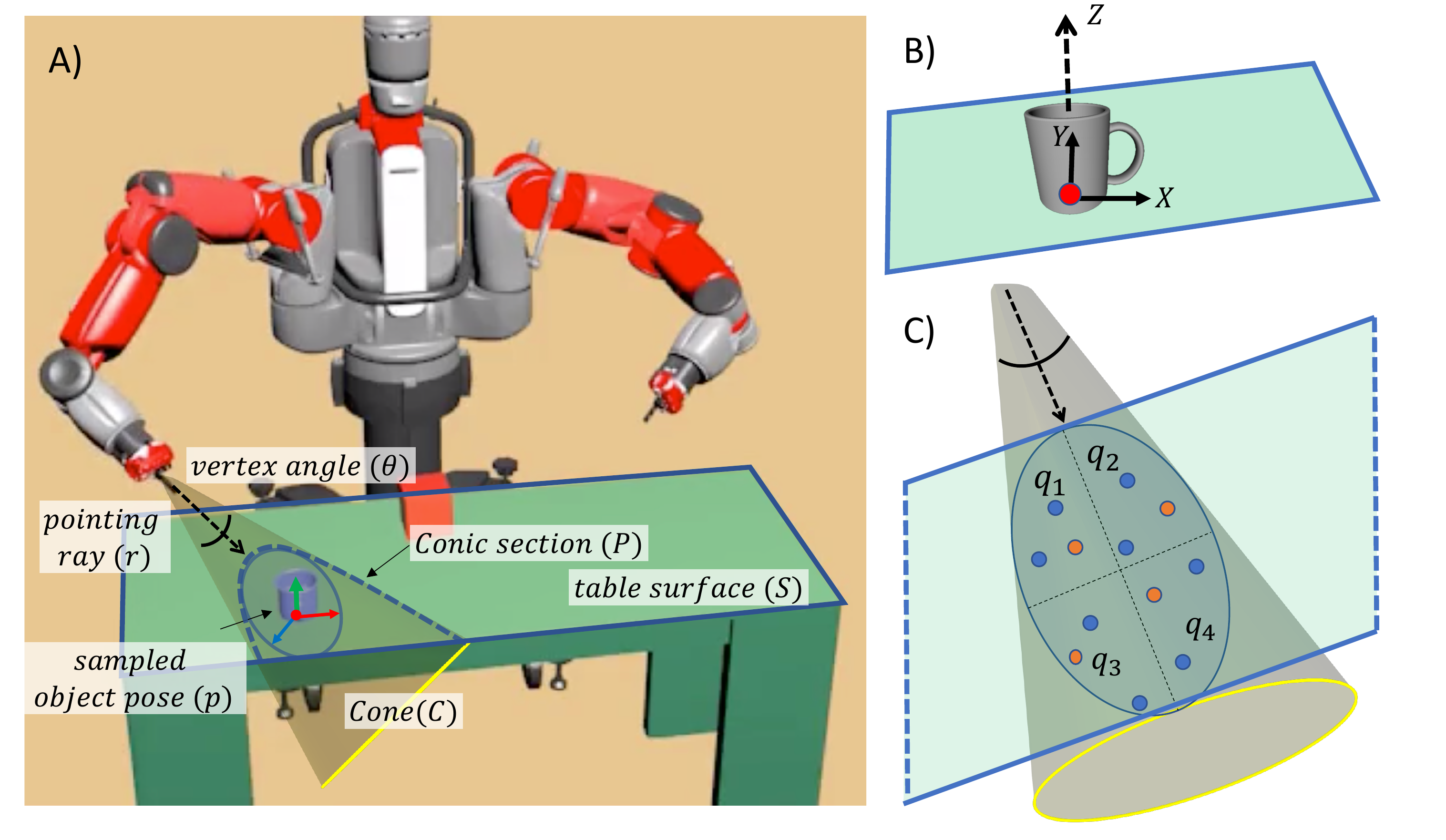}
\caption{(A) Workspace setup showing the pointing cone and the corresponding conic section on the table. (B) The degrees-of-freedom considered for placement of the object on the table. (C) Sampling policy to sample object poses within the conic section.}
    \label{fig:pointing}
\end{figure}

\noindent\textbf{Pointing Action}: Within its reachable workspace the end-effector of the manipulator can attain different orientations to fully specify a reachable \textit{pose} $p$, which describes its position and orientation.  The robots we study have a directional tooltip that viewers naturally see as projecting a ray $r$ along its axis outward into the scene.  In understanding pointing as communication, the key question is the relationship between the ray $r$ and the spatial values $\xinit$ and $\xfinal$ that define the pick-and-place task.

To make this concrete, we distinguish between the \emph{target} of pointing and the \emph{intent} of pointing. Given the ray $r$ coming out of the end-effector geometry, we define the target of the pointing as the intersection of this ray on the stable surface, $$x^*= r\cap S.$$ Meanwhile, the intent of pointing specifies one component of a pick-and-place task.  There are two cases:
\begin{itemize}
    \item [-] \textit{Referential Pointing:} The pointing action is intended to identify a target object $o$ to be picked up. This object is the \textit{referent} of such an action. We can find $\xinit$, based on the present position of $o$.
    \item [-] \textit{Locating Pointing:} The pointing action is intended to identify the location in the workspace where the object needs to be placed, i.e, $\xfinal$.
\end{itemize}

We study effective ways to express intent for a pick-and-place task. In other words, what is the relationship between a pointing ray $r$ and the location $\xinit$ or $\xfinal$ that it is intended to identify?  To assess these relationships, we ask human observers to view animations expressing pick-and-place tasks and classify their interpretations.  To understand the factors involved, we investigate a range of experimental conditions.

\section{Experiments}
\label{experiments}

Our experiments share a common animation platform, described in the Experimental Setup, and a common Data Collection protocol.  The experiments differ in presenting subjects with a range of experimental conditions, as described in the corresponding section.  All of the experiments described here together with the methods chosen to analyze the data were based on a private but approved pre-registration on \textit{aspredicted.org}. The document is publicly available at: \url{https://aspredicted.org/cg753.pdf}.

\subsection{Experiment Setup}
Each animation shows a simulated robot producing two pointing gestures to specify a pick-and-place task.  Following the animation, viewers are asked whether a specific image represents a possible result of the specified task.\\

\noindent\textbf{Robotic Platforms} The experiments were performed on two different robotic geometries, based on a \textit{Rethink Baxter}, and a \textit{Kuka IIWA14}.  The \textit{Baxter} is a dual-arm manipulator with two arms mounted on either side of a static torso. The experiments only move the right arm of the \textit{Baxter}. The \textit{Kuka}  consists of a single arm that is vertically mounted, i.e., points upward at the base. In the experiments the robots are shown with a singly fingered tool-tip, where the pointing ray is modeled as the direction of this tool-tip.\\

\begin{figure}[t]
    \centering
    \includegraphics[width=0.45\textwidth]{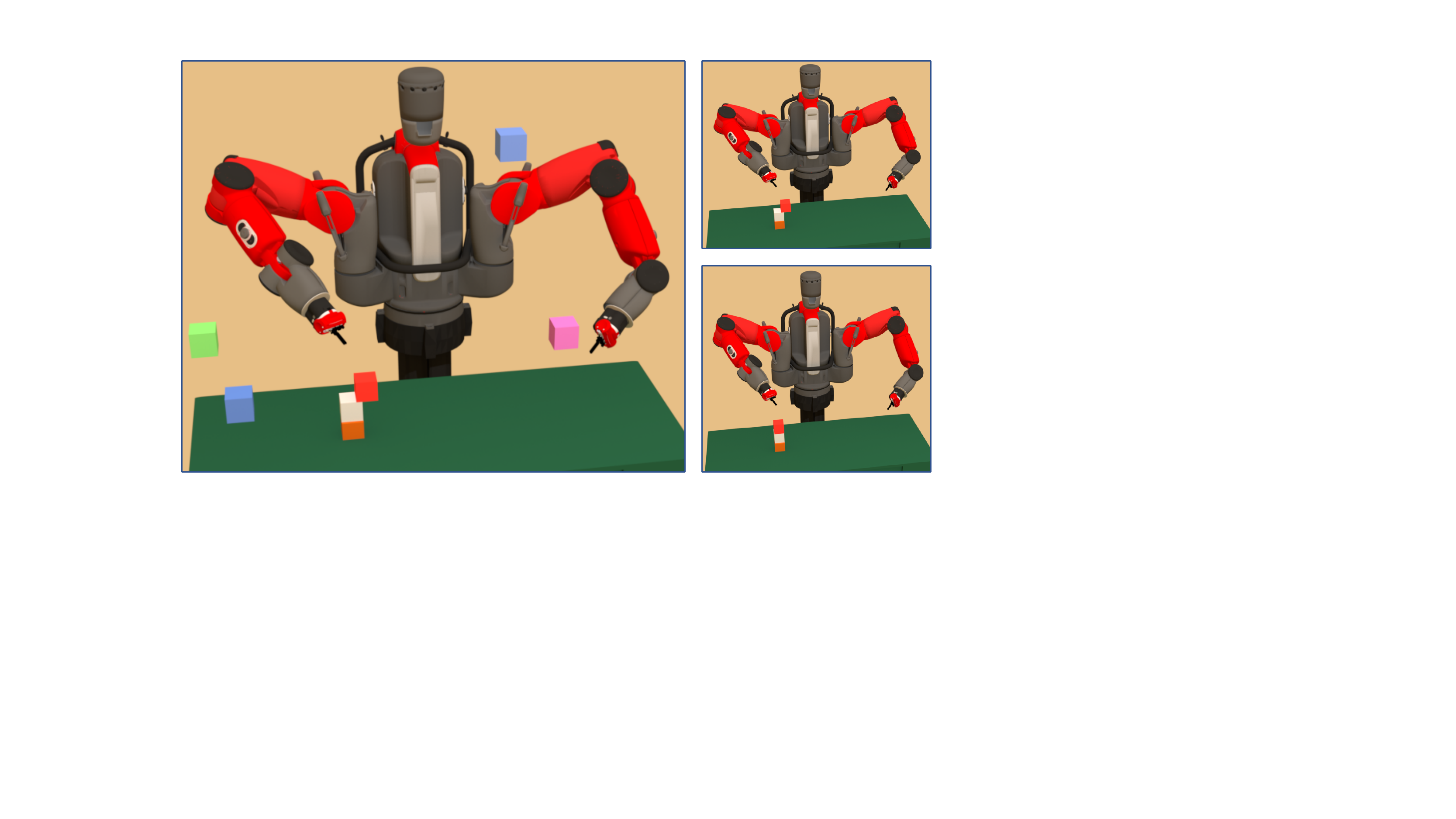}
    \caption{(Left) In an unnatural scene, a gesture pointing to an unstable position (edge of the stack) is deemed correct. (Right) In natural scenes, although the robot points to the edge of the stack, a physically realistic object position gets more user vote than the unstable position.}
    \label{fig:natural}
\end{figure}

\noindent\textit{Note} The real Baxter robot possesses a heads-up display that can be likened to a `head'. This has been removed in the simulations that were used in this study (as shown for example in Figure~\ref{fig:natural}).\\

\noindent\textbf{Workspace Setup} Objects are placed in front of the manipulators. In certain trials a table is placed in front of the robot as well, and the objects rest in stable configurations on top of the table. A pick-and-place task is provided specified in terms of the positions of one of the objects. \\

\noindent\textbf{Objects}  The objects used in the study include small household items like mugs, saucers and boxes (cuboids), that are all placed in front of the robots.\\

\noindent\textbf{Motion Generation}  The end-effector of the manipulator is instructed to move to pre-specified waypoints, designed for the possibility of effective communication, that typically lie between the base of the manipulator and the object itself. Such waypoints fully specify both the position and orientation of the end-effector to satisfy \textit{pointing actions}. The motions are performed by solving Inverse Kinematics for the end-effector geometry and moving the manipulator along these waypoints using a robotic motion planning library \cite{pracsys2014}. The motions were replayed on the model of the robot, and rendered in \textit{Blender}.\\

\begin{figure}[t]
    \centering
    \includegraphics[width=0.48\textwidth]{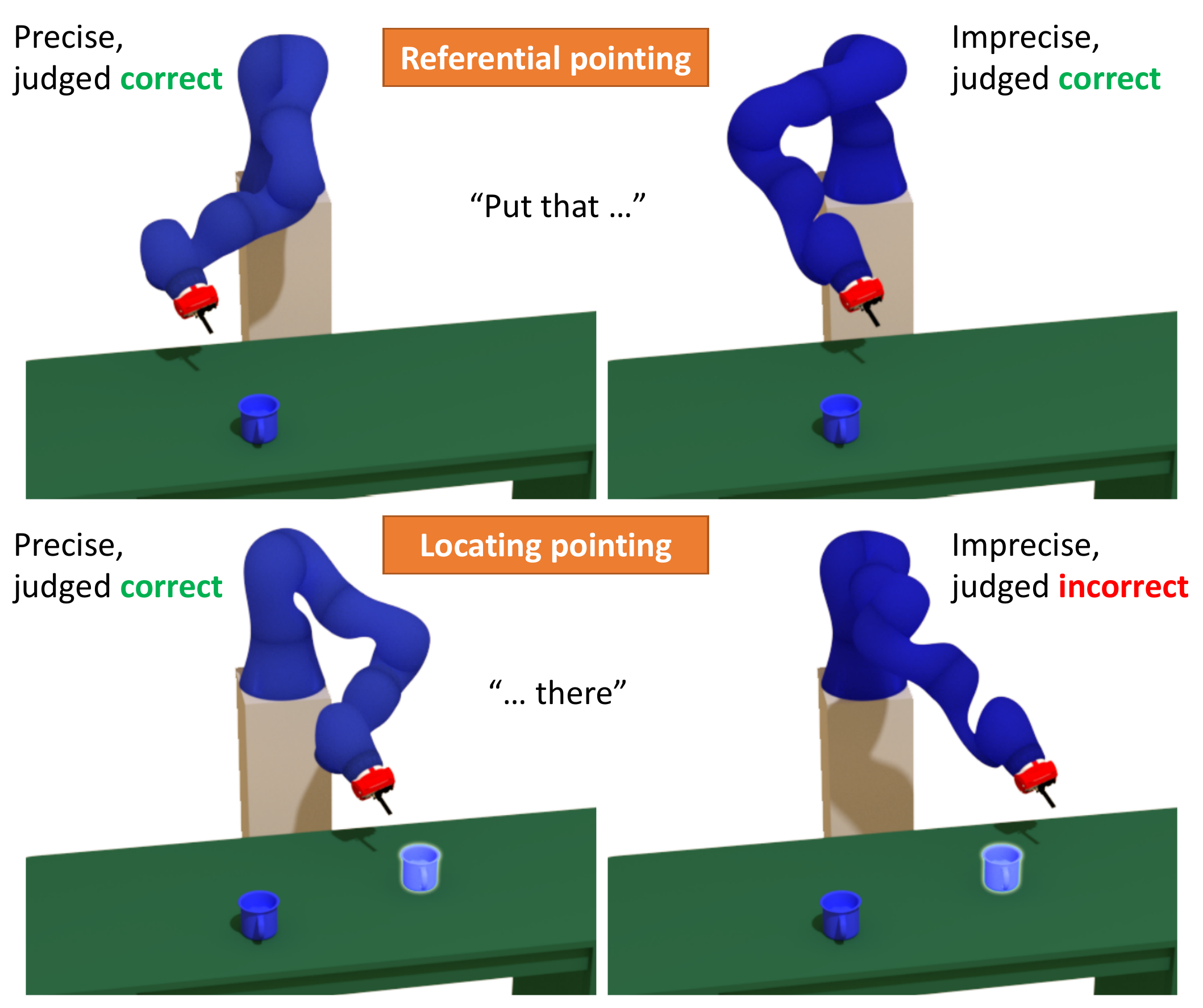}
    \caption{The image shows the differences between referential (\textit{top}) and locating pointing (\textit{bottom}), demonstrated on a robotic manipulator, \textit{Kuka IIWA14}. An overlay of the object is shown at the placement location where locating pointing needs to be directed.  Human subjects are more flexible in the interpretation of  referential pointing than with that of locating pointing.}
    \label{fig:spatial}
\end{figure}

\noindent\textbf{Pointing Action Generation} Potential pointing targets are placed using a cone $C(r, \theta)$, where $r$ represents the pointing ray and $\theta$ represents the vertex angle of the cone. As illustrated in Fig~\ref{fig:pointing}, the cone allows us to assess the possible divergence between the pointing ray and the actual location of potential target objects on the rest surface $S$. 

Given a pointing ray $r$, we assess the resolution of the pointing gesture by sampling $N$ object poses $p_i, i=1:N$ in $P=C(r, \theta) \cap S$---the intersection of the pointing cone with the rest surface.  While $p_i$ is the 6d pose for the object with translation $t \in R^3$ and orientation $R \in SO(3)$ only 2  degrees-of-freedom $(x, y)$ corresponding to $t$ are varied in the experiments. By fixing the $z$ coordinate for translation and restricting the z-axis of rotation to be perpendicular to $S$, it is ensured that the object rests in a physically stable configuration on the table.

The $N$ object poses are sampled by fitting an ellipse within $P$ and dividing the ellipse into 4 quadrants $q_1\ldots q_4$ (See Figure~\ref{fig:pointing} (C)). Within each quadrant $q_i$ the $N/4$ $(x,y)$ positions are sampled uniformly at random. For certain experiments additional samples are generated with an objective to increase coverage of samples within the ellipse by utilizing a dispersion measure.\\

\noindent\textbf{Speech} Some experiments also included verbal cues with phrases like `\textit{Put that there}' along with the pointing actions. It was very important for the pointing actions and these verbal cues to be in synchronization. To fulfill this we generate the voice using Amazon Polly with text written in SSML format and make sure that peak of the gesture (the moment a gesture comes to a stop) is in alignment with the peak of each audio phrase in the accompanying speech. During the generation of the video itself we took note of the peak moments of the gestures and then manipulated the duration between peaks of the audio using SSML to match them with gesture peaks after analyzing the audio with the open-source tool PRAAT (www.praat.org).

\subsection{Data Collection}

Data collection was performed in \textit{Amazon Mechanical Turk}. All subjects agreed to a consent form and were compensated at an estimated rate of \textit{USD 20} an hour. The subject-pool was restricted to non-colorblind US citizens. Subjects are presented a rendered video of the simulation where the robot performs one referential pointing action, and one locating pointing action which amounts to it pointing to an object, and then to a final location. During these executions synchronized speech is included in some of the trials to provide verbal cues.

Then on the same page, subjects see the image that shows the result of the pointing action. They are asked whether the result is (a) correct, (b) incorrect, or (c) ambiguous.  

To test our hypothesis, we studied the interpretation of the two pointing behaviors in different contexts. Assuming our conjecture and a significance level of 0.05, a sample of 28 people in each condition is enough to detect our effect with a 95\% power.  Participants are asked to report judgments on the interpretation of the pointing action in each class.  Each participant undertakes two trials from each class.  The range of different cases are described below.  Overall, the data collection in this study involved over 7,290 responses to robot pointing actions.\footnote{ The data, code, and videos are available at  \url{https://github.com/malihealikhani/That_and_There}.}

\subsection{Experimental Conditions}

We used our experiment setup to generate videos and images from the simulation for a range of different conditions.

\paragraph{Referential vs Locating}
In this condition, to reduce the chances of possible ambiguities, we place only one mug is on the table. The \textit{Baxter} robot points its right arm to the mug and then points to its final position, accompanied by a synchronized verbal cue, \textit{``Put that there.''}

We keep the motion identical across all the trials in this method. 
We introduce a variability in the initial position of the mug by sampling $8$ random positions within conic sections subtending $45^{\circ} , 67.5^{\circ}, $ and $90^{\circ}$ on the surface of the table. New videos are generated for each such position of the mug.
This way we can measure how flexible subjects are to the variation of the initial location of the referent object. 

To test the effect for the locating pointing action, we test similarly sampled positions around the final pointed location, and display these realizations of the mug as the result images to subjects, while the initial position of the mug is kept perfectly situated. 

A red cube that is in the gesture space of the robot, and is about twice as big as the mug is placed on the other side of the table as a visual guide for the subjects to see how objects can be placed on the table. We remove the tablet that is attached to Baxter's head for our experiments.\\ 

\noindent\textit{Effect of speech} In order to test the effect of speech on the disparity between the kinds of pointing actions, a set of experiments were designed under the \textit{Referential vs Locating} method with and without any speech. All subsequent methods will include verbal cues during their action execution. These cues are audible in the video.

\paragraph{Reverse Task} One set of experiments are run for the pick-and-place task with the initial and final positions of the object flipped during the reverse task. As opposed to the first set of experiments, the robot now begins by pointing to an object in the middle of the table, and then to an area areas towards the table's edge, i.e., the pick and place positions of the object are `reversed'. 

The trials are meant to measure the sensitivity of the subjects in pick trials to the direction of the pointing gestures and to the absolute locations that the subjects thought the robot was pointing at.

This condition is designed to be identical to the basic Referential vs Locating study, except for the direction of the action. The motions are still executed on the \textit{Baxter's} right arm. 

\paragraph{Different Robotic Arm}
In order to ensure that the results obtained in this study are not dependent on the choice of the robotic platform or its visual appearance, a second robot---a singly armed industrial \textit{Kuka} manipulator---is also evaluated in a Referential vs Locating study (shown in Figure~\ref{fig:spatial}).

\begin{figure}[t]
    \centering
    \includegraphics[width=0.40\textwidth]{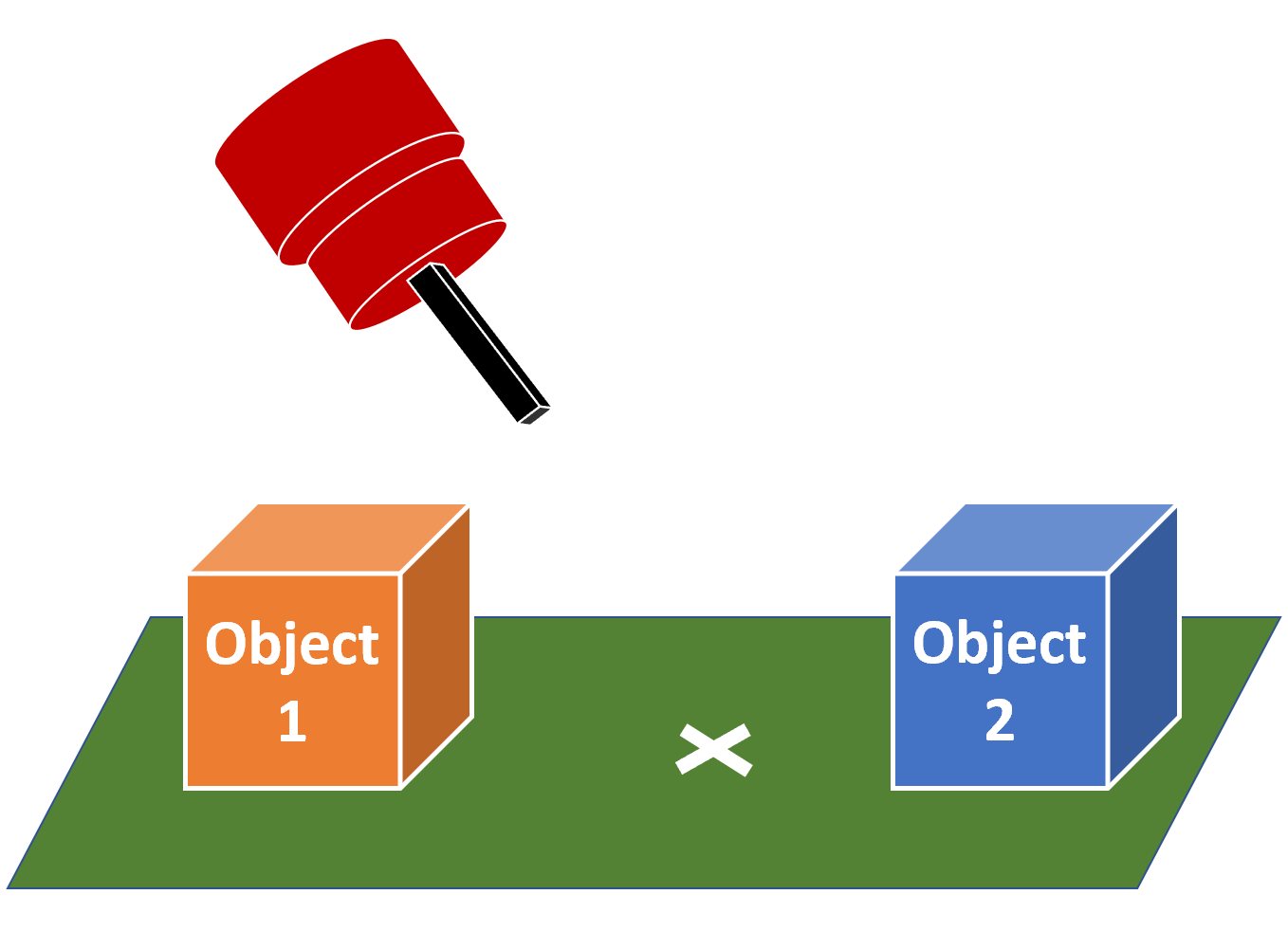}
    \caption{A cluttered trial consists of collecting the response from a human subject when the position of the referential pointing action lies between two objects.}
    \label{fig:cluttered_trial}
\end{figure}

\paragraph{Cluttered Scene}
To study how the presence of other objects would change the behavior of referential pointing, we examine the interpretation of the pointing actions when there is more than one \textit{mug} on the table. Given the instructions to the subjects, both objects are candidate targets. This experiment allows the investigation of the effect of a distractor object in the scene on referential pointing.  

We start with a setup where there are two mugs placed on the table (similar to the setup in Figure~\ref{fig:cluttered_trial}). One is a target mug placed at position $\xobject$ and a distractor mug at position $\xdistractor$. With the robot performing an initial pointing action to a position $\xinit$ on the table. Both the objects are sampled around $\xinit$ along the diametric line of the conic section arising from increasing cone angles of $45^\circ, 67.5^\circ, $ and $90^\circ$, where the separation of $\xobject$, and $\xdistractor$ is equal to the length of the diameter of the conic section, $D$. The objects are then positioned on the diametric line with a random offset between $[-\frac{D}{2}, \frac{D}{2}]$ around $\xinit$ and along the line. This means that the objects are at various distances apart, and depending upon the offset, one of the objects is nearer to the pointing action. The setup induces that the nearer mug serves as the \textit{object}, and the farther one serves as the \textit{distractor}. The motions are performed on the \textit{Baxter's} right arm. The camera perspective in simulation is set to be facing into the pointing direction. The subjects in this trial are shown images of the instant of the referential pointing action.

\begin{figure}[h!]
    \centering
    \includegraphics[width=0.47\textwidth] {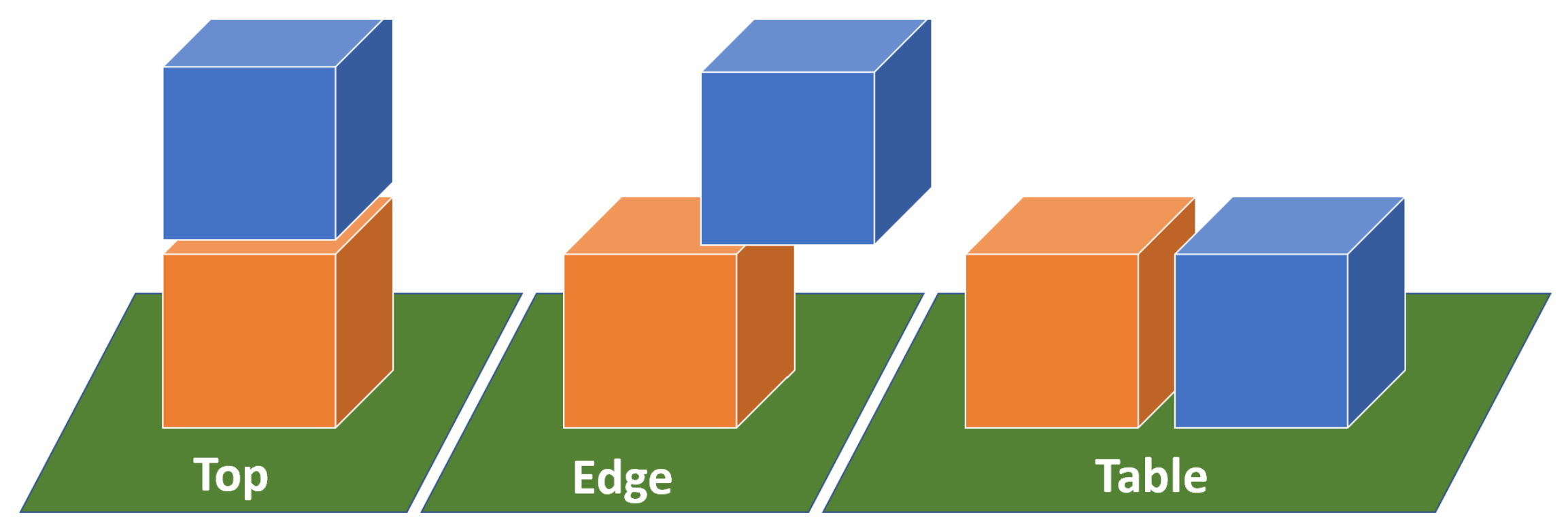}
    \caption{
    The diagram shows the three different configurations of the placement of a blue cuboid object evaluated in the \textit{Natural vs Unnatural} trials. 
    }
    \label{fig:topedgetable}
\end{figure}

\paragraph{Natural vs Unnatural scene}
In this condition we study how the contextual and physical understanding of the world impacts the interpretation of pointing gestures. We generate a scenario for locating pointing in which the right arm of the \textit{Baxter} points to a final placement position for the cuboidal object on top of a stack of cuboidal objects but towards the edge which makes it physically unstable. The final configurations of the object (Figure~\ref{fig:topedgetable}) shown to the users were a) object lying on top of the stack b) object in the unstable configuration towards the edge of the stack and c) object at the bottom of the stack towards one side. New videos are generated for each scenario along with verbal cues.

The pointing action, as well as the objects of interest, stay the identical between the natural, and unnatural trials. The difference lies in other objects in the scene that could defy gravity and float in the unnatural trials. The subjects were given a text-based instruction at the beginning of an unnatural trial saying they were seeing a scene where ``gravity does not exist.''

\paragraph{Different verbs}  
To test if the effect is specific to the verb \textit{put}, we designed a control condition where everything remained the same as the Referential vs Locating trials except the verb \textit{put} which we replaced with \textit{place, move} and \textit{push}. Here again we collect 30 data points for each sampled $x^*$.

\section{Analysis}
\label{analysis}

\begin{figure*}[ht!]

    \centering
    \includegraphics[width=0.5\textwidth] {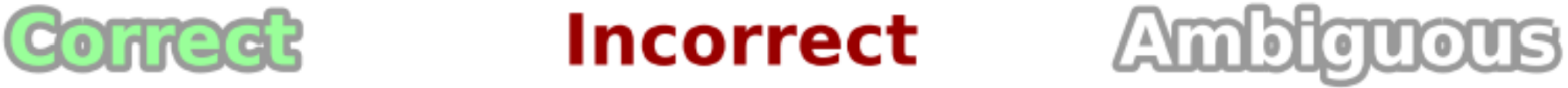}\\
    \includegraphics[width=0.325\textwidth ] {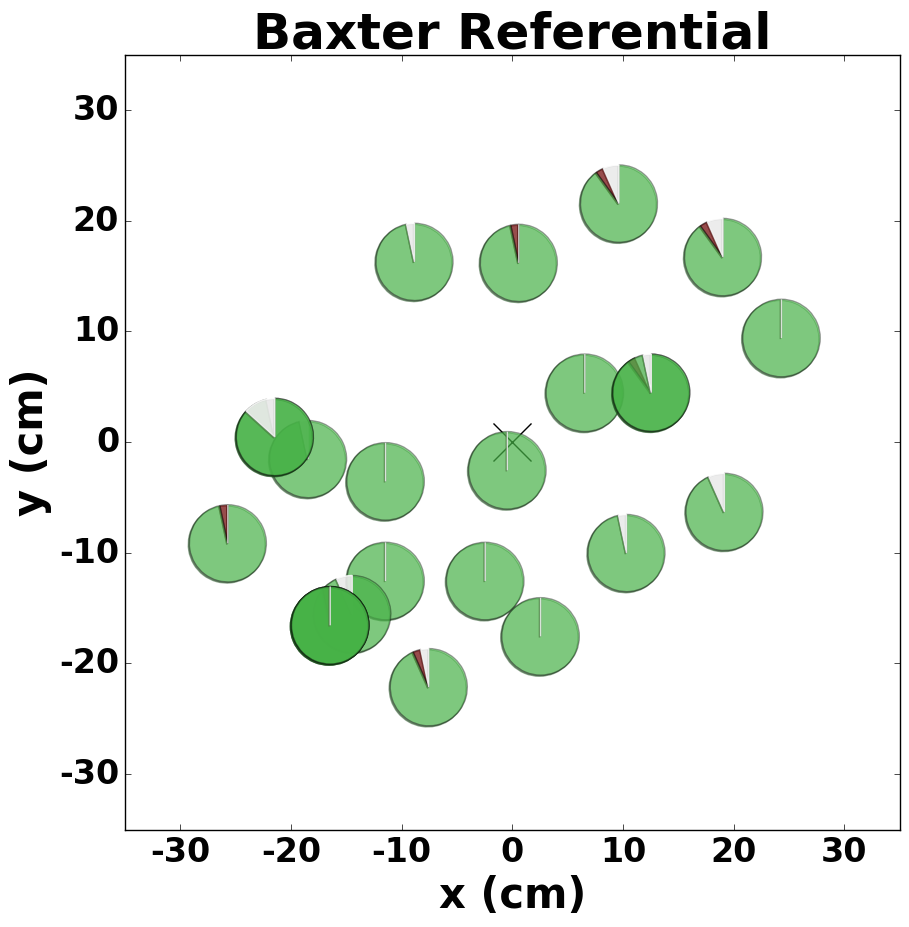}
    \includegraphics[width=0.325\textwidth ] {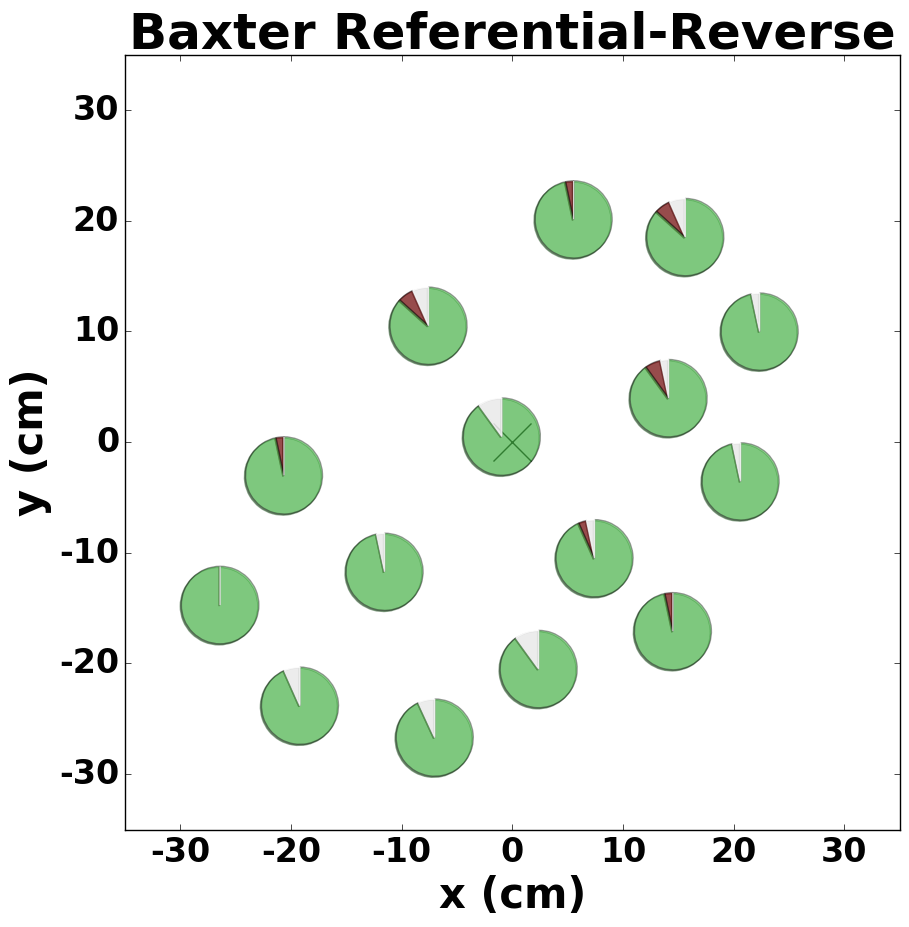}
    \includegraphics[width=0.325\textwidth ]{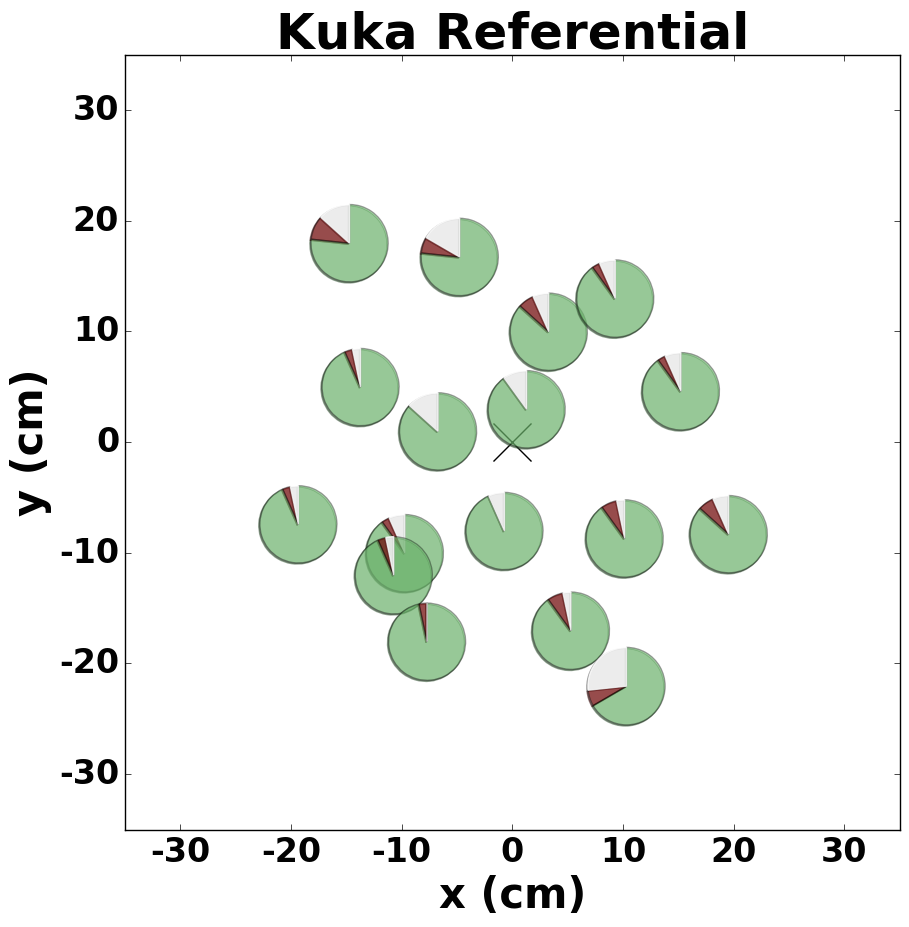}
    \includegraphics[width=0.325\textwidth ]{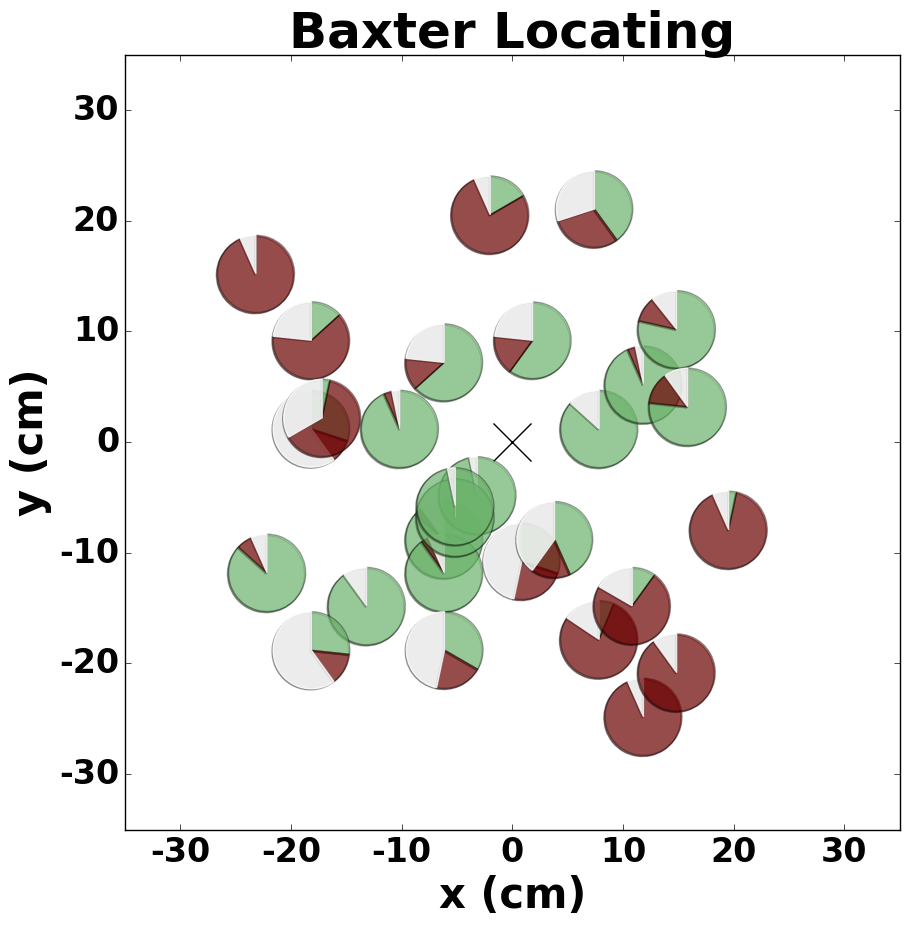}
    \includegraphics[width=0.325\textwidth ]{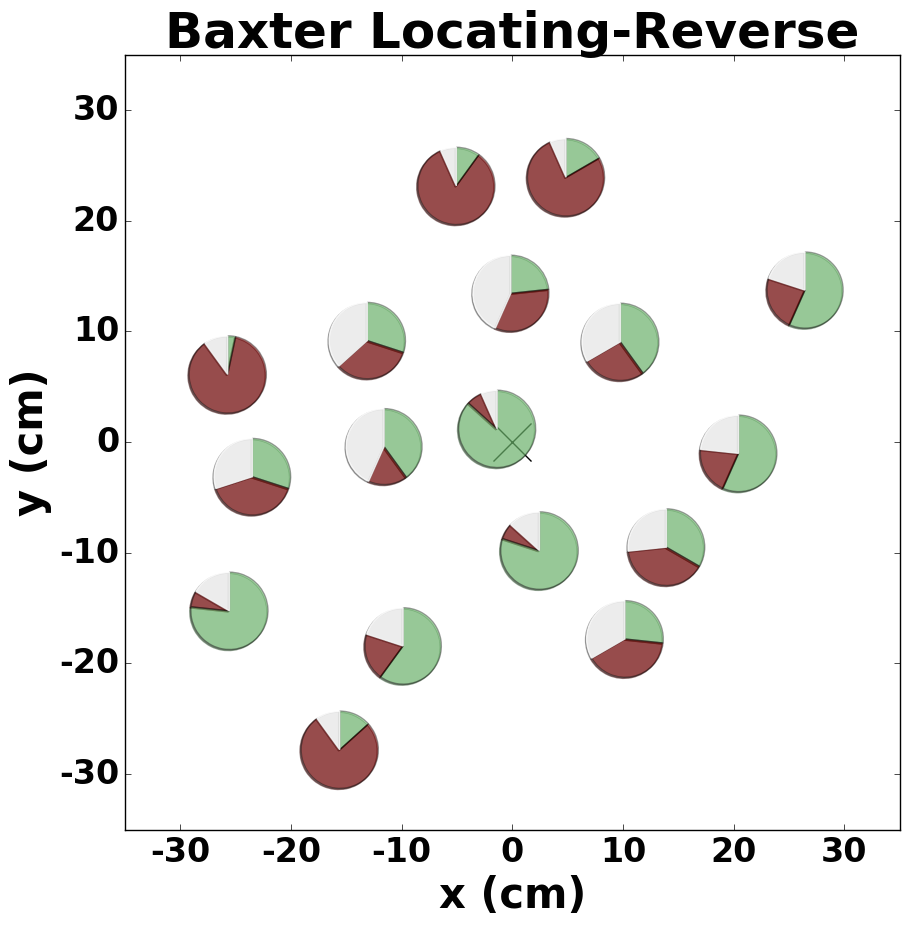}
    \includegraphics[width=0.325\textwidth ]{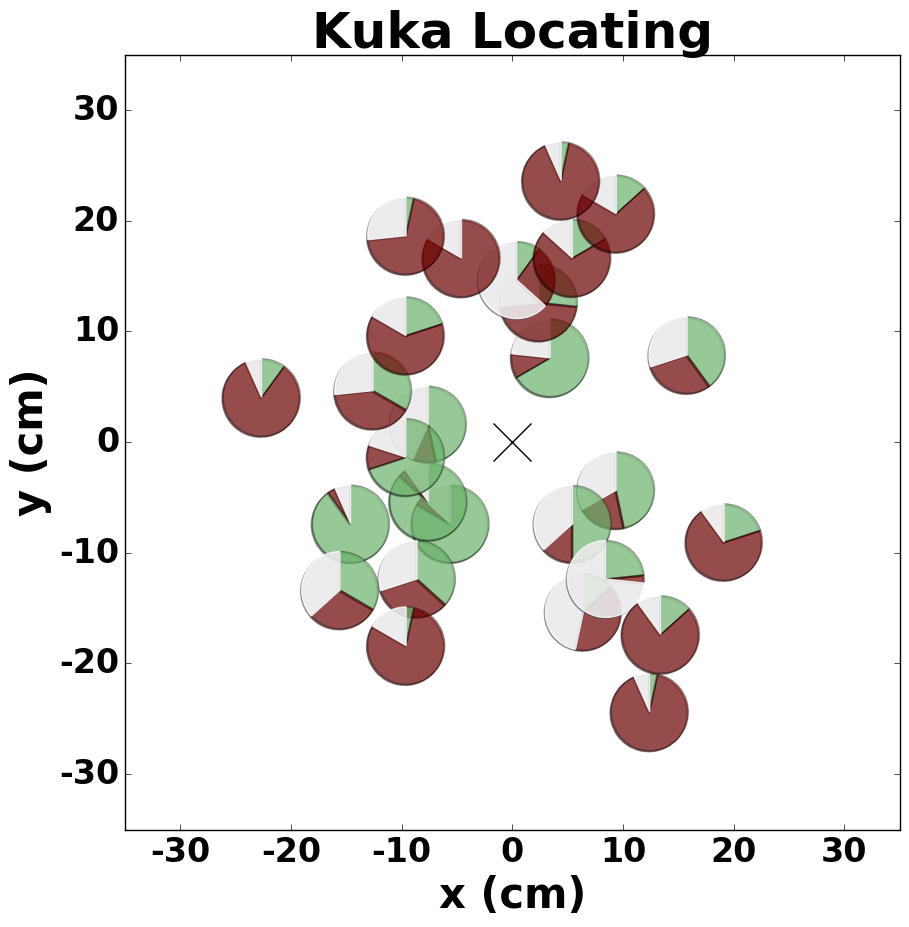}
    \caption{The aggregated results from the referential versus spatial trials for the \textit{Baxter} and \textit{Kuka} robots. The locations of the responses correspond to the center of the circles, and are plotted in the coordinate frame centered at the position of the pointing action, marked with $\times$. The circles show the fraction of correct (grey), incorrect (black) and ambiguous (white) responses.}
    \label{fig:aggregatesimple}
\end{figure*}

\paragraph{Referential vs Locating}
We study how varying the target of the pointing action from a referent object to a part of the space changes the interpretation of the pointing action by comparing the interpretation of the position of the pointing action $x^*$ in each condition. 

Figure~\ref{fig:aggregatesimple} shows the results of the experiment. The plot shows the spread of \textit{correct, incorrect, ambiguous} responses over the sampled positions about the location of referential vs locating pointing actions. The referential data demonstrates the robustness of the interpretation. Most of the responses were overwhelmingly \textit{correct}, for both robots, in interpreting a referent object in the \textit{pick} part of a pick-and-place task. The locating pointing shows a much higher sensitivity to an accuracy of $x^*$ with respect to the true final placement. This comes up as a larger incidence of \textit{incorrect} and \textit{ambiguous} responses from the human subjects. This trend is true for the reverse trial as well.

While the study attempts to separate out and measure the critical aspects of the interpretation of robotic pointing actions some ambiguities like those arising out of perspective of the camera being projected onto a simulated 2D video or image are unavoidable. We suspect that the observed stretch of \textit{correct} responses in spatial trials is due to perspective.

To test our hypothesis that Referential pointing is interpreted less precisely than Locating pointing we performed a Chi-squared test and compared the proportion of \textit{correct}, \textit{incorrect} and \textit{ambiguous} responses in referential and spatial trials. The results of the test shows that these two classes are statistically significantly different ($\chi^2= 13.89, p = 0.00096$).

To study if we are observing the same effects in the results of the reverse trial, no speech trial and the Kuka trial, we ran an equivalence test following the two one-sided tests method as described in \cite{lakens2017equivalence}, where each test is a pooled $z$-test with no continuity
correction with a significance level of 0.05. We found changing the robot, removing the speech and changing the direction of the pointing action to make no difference in the interpretation of locating pointing and referential pointing within any margin that is less than 5\%.

\begin{table}[h]
\label{tab:naturaltrial}
\begin{tabular}{lllll}
          &               & correct     & incorrect & ambiguous \\ \hline
unnatural & top           & 12          & 9         & 9         \\
          & \textbf{edge} & \textbf{24} & 2         & 4         \\
          & table         & 2           & 2         & 26        \\ \hline
natural   & \textbf{top}  & \textbf{26} & 3         & 1         \\
          & edge          & 9           & 11        & 10        \\
          & table         & 7           & 13        & 12        \\ \hline
\end{tabular}
\caption{Results of the unnatural scene and natural scene. (Numbers are out of 30.)}
\label{tab:natural-unnatural}
\end{table}

\paragraph{Natural vs Unnatural}

As shown in Table~\ref{tab:natural-unnatural} we observed in the natural scene, when the end-effector points towards the edge of the cube that is on top of the stack, subjects place the new cube on top of the stack or on the table instead of the edge of the cube. However, in the unnatural scene, when we explain to subjects that there is no gravity, a majority agree with the final image that has the cube on the \textit{edge}. To test if this difference is statistically significant, we use the Fisher exact test \cite{10.2307/2340manuelli2019kpam521}. 
The test statistic value is $0.0478$. The result is significant at $p < 0.05$.

\paragraph{Different verbs}
The results of the Chi-squared test shows that in spatial trials when we replace \textit{put} with \textit{place}, \textit{push} and \textit{move}, the differences of the distributions of \textit{correct}, \textit{incorrect} and \textit{ambiguous} responses are not statistically significant ($\chi=0.2344 $, $p = 0.971$). The coefficients of the multinomial logistic regression model and the $p$-values also suggest that the differences in judgements 
with different verbs are not statically significant ($b<0.0001$ , $p>0.98$).

\begin{figure}[ht]
    \centering
    \includegraphics[width=\linewidth]{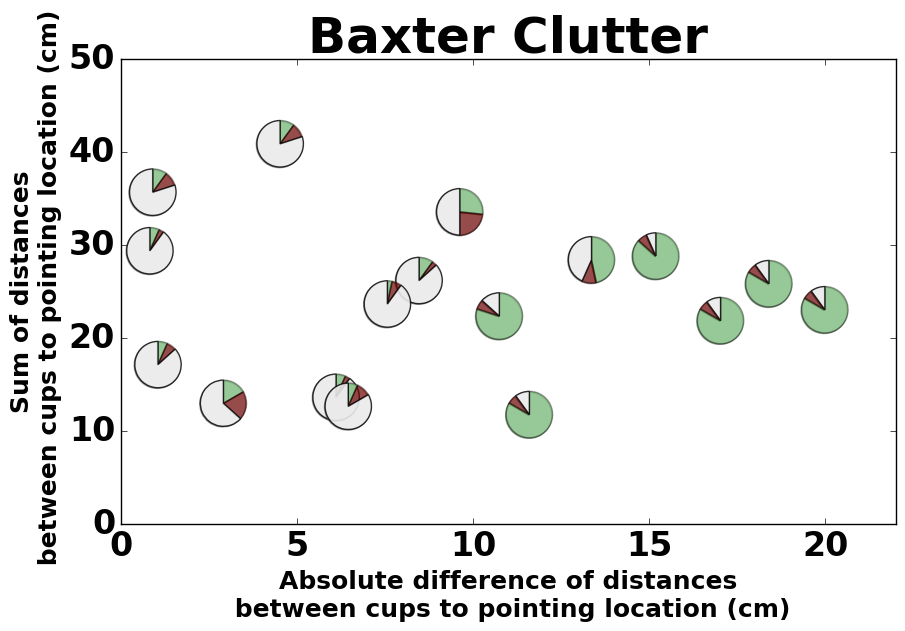}
    \caption{The scatter plot represents the spread of responses where human subjects chose the \textit{nearer cup} (green), \textit{farther} cup (red), and ambiguous (white). The \textit{x-axis} represents the absolute difference between the distances of each cup to the locations of pointing, the \textit{y-axis} represents the total distance between the two cups.}
    \label{fig:cluttered}
\end{figure}
\paragraph{Cluttered}
The data from these trials show how human subjects select between the two candidate target objects on the table. Since the instructions do not serve to disambiguate the target mug, the collected data show what the observers deemed as the \textit{correct} target.  Figure~\ref{fig:cluttered} visualizes subjects' responses across trials.  The location of each pie uses the $x$-axis to show how much closer one candidate object is to the pointing target than the other, and uses the $y$-axis to show the overall imprecision of pointing.  Each pie in Figure~\ref{fig:cluttered} shows the fraction of responses across trials that recorded the nearer (green) mug as correct compared to the farther mug (red). The white shaded fractions of the pies show the fraction of responses where subjects found the gesture ambiguous.

As we can see in Figure~\ref{fig:cluttered}, once the two objects are roughly equidistant the cups from the center of pointing (within about 10cm), subjects tend to regard the pointing gesture as ambiguous, but as this distance increases, subjects are increasingly likely to prefer the closer target.  In all cases, wherever subjects have a preference for one object over the other, they subjects picked the mug that was the nearer target of the pointing action more often than the further one.

\section{Human Evaluation of Instructions}

After designing and conducting our experiments, we became concerned that subjects might regard imprecise referential pointing as understandable but unnatural.  If they did, their judgments might combine ordinary interpretive reasoning with additional effort, self-consciousness or repair.  We therefore added a separate evaluation to assess how natural the generated pointing actions and instructions are. We recruited 480 subjects from Mechanical Turk using the same protocol described in our Data Collection procedure, and asked them to rank how natural they regarded the instruction on a scale of \textit{0 to 5}. 

The examples were randomly sampled from the videos of the referential pointing trials that we showed to subjects for both the Baxter and Kuka robots. These examples were selected in a way that we obtained equal number of samples from each cone. The average rating for samples from the $\ang{45}$, $\ang{67.5}$ and $\ang{90}$ cone are $3.625, 3.521$
and $3.650$ respectively. For Kuka, the average rating for samples from the $\ang{45}$, $\ang{67.5}$ and $\ang{90}$ cone are $3.450, 3.375$, and $3.400$. Overall, the average for Baxter is $3.600$, and for Kuka is $3.408$. The differences between Kuka and Baxter and the differences across cones are not statistically significant ($t \leq |1.07|, p > 0.1 $).  Thus we have no evidence that subjects regard imprecise pointing as problematic.

\section{Design Principles}

The results of the experiments suggest that locating pointing is interpreted rather precisely, where referential pointing is interpreted relatively flexibly.  This naturally aligns with the possibility for alternative interpretations.  For spatial reference, any location is a potential target.  By contrast, for referential pointing, it suffices to distinguish the target object from its distractors.

We can characterize this interpretive process in formal terms by drawing on observations from the philosophical and computational literature on vagueness \cite{devault2004interpreting,graff2000shifting,kyburg2000fitting}.  Any pointing gesture starts from a set of candidate interpretations $D \subset \mathcal{W}$ determined by the context and the communicative goal.  In unconstrained situations, locating pointing allows a full set of candidates $D = \mathcal{W}.$  If factors like common-sense physics impose task constraints, that translates to restrictions on feasible targets $CS$, leading to a more restricted set of candidates $D = CS \cap \mathcal{W}$.  Finally, for referential pointing, the potential targets are located at $x_1 \ldots x_N \in S$, and $D = \{ x_1 \ldots x_N \}.$

Based on the communicative setting, we know that the pointing gesture, like any vague referring expression, must select at least one of the possible interpretations \cite{kyburg2000fitting}.  We can find the best interpretation by its distance to the target $x^*$ of the pointing gesture.  Using $d(x,x^*)$ to denote this distance, gives us a threshold $$\theta = \min_{x \in D} d(x, x^*).$$

Vague descriptions can't be sensitive to fine distinctions \cite{graff2000shifting}.  So if a referent at $\theta$ is close enough to the pointing target, then another at $\theta + \epsilon$ must be close enough as well, for any value of $\epsilon$ that is not significant in the conversational context. Our results suggest that viewers regard 10cm (in the scale of the model simulation) as an approximate threshold for a significant difference in our experiments.

In all, we predict that a pointing gesture is interpreted as referring to $\{x \in D | d(x,x^*) \leq \theta + \epsilon\}.$  We explain the different interpretations through the different choice of $D$.

\paragraph{Locating Pointing}  For unconstrained locating pointing, $x^* \in D$, so $\theta=0$.  That means, the intended placement cannot differ significantly from the pointing target.  Taking into account common sense, we allow for small divergence that connects the pointing, for example, to the closest stable placement.

\paragraph{Referential Pointing}  For referential pointing, candidates play a much stronger role.  A pointing gesture always has the closest object to the pointing target as a possible referent.  However, ambiguities arise when the geometries of more than one object intersect with the $\theta+\epsilon$-neighborhood of $x^*$.   We can think of that, intuitively, in terms of the effects of $\theta$ and $\epsilon$.  Alternative referents give rise to ambiguity not only when they are too close to the target location ($\theta$) but even when they are simply not significantly further away from the target location ($\epsilon$).  

\section{Conclusion and Future Work}
\label{conclusion}

We have presented an empirical study of the interpretation of simulated robots instructing pick-and-place tasks.  Our results show that robots can effectively combine pointing gestures and spoken instructions to communicate both object and spatial information. We offer an empirical characterization---the first, to the best of the authors' knowledge---of the use of robot gestures to communicate precise spatial locations for placement purposes.  We have suggested that pointing, in line with other vague references, give rise to a set of candidate interpretations that depend on the task, context and communicative goal.  Users pick the interpretations that are not significantly further from the pointing ray than the best ones.  This contrasts with previous models that required pointing gestures to target a referent exactly or fall within a context-independent pointing cone.

Our work has a number of limitations that suggest avenues for future work.   It remains to implement the design principles on robot hardware, explore the algorithmic process for generating imprecise but interpretable gestures, and verify the interpretations of physically co-present viewers.  Note that we used a 2D interface, which can introduce artifacts, for example from the effect of perspective.  In addition, robots can in general trade off pointing gestures with other descriptive material in offering instructions.  Future work is needed to assess how such trade-offs play out in location reference, not just in object reference.

More tight-knit collaborative scenarios need to be explored, including ones where multiple pick-and-place tasks can be composed to communicate more complex challenges and ones where they involve richer human environments.  Our study of  \textit{common sense} settings opens up intriguing avenues for such research, since it suggests ways to take into account background knowledge and expectations to narrow down the domain of possible problem specifications in composite tasks like \textit{``setting up a dining table.''} 

While the current work studies the modalities of pointing and verbal cues, effects of including additional robotic communication in the form of heads-up displays or simulated eye-gaze would be other directions to explore. Such extensions would require lab experiments with human subjects and a real robot.  This is the natural next step of our work. 

\section{Acknowledgments}
The research presented here is supported by NSF Awards IIS-1526723, IIS-1734492, IIS-1723869 and CCF-1934924. Thanks to the anonymous reviewers for helpful comments. We would also like to thank the Mechanical Turk participants for their contributions.


\bibliographystyle{aaai}
\end{document}